\documentclass[conference]{IEEEtran}
\IEEEoverridecommandlockouts
\usepackage{cite}
\usepackage{amsmath,amssymb,amsfonts}
\usepackage{algorithmic}
\usepackage{hyperref}
\usepackage[capitalise,nameinlink,noabbrev]{cleveref}
\usepackage{graphicx}
\usepackage{textcomp}
\usepackage{xcolor}
\usepackage{booktabs}
\definecolor{ForestGreen}{rgb}{0.21,0.88,0.56}
\usepackage{multirow}
\def\BibTeX{{\rm B\kern-.05em{\sc i\kern-.025em b}\kern-.08em
    T\kern-.1667em\lower.7ex\hbox{E}\kern-.125emX}}
\begin{document}

\title{GaussianTrimmer: Online Trimming Boundaries for 3DGS Segmentation}

\author{Liwei Liao\\
Peking University Shenzhen Graduate School\\
{\tt\small levio@pku.edu.cn}
\and
Ronggang Wang\\
Peking University Shenzhen Graduate School\\
{\tt\small rgwang@pkusz.edu.cn}}

\maketitle

\begin{abstract}
With the widespread application of 3D Gaussians in 3D scene representation, 3D scene segmentation methods based on 3D Gaussians have also gradually emerged. However, existing 3D Gaussian segmentation methods basically segment on the basis of Gaussian primitives. Due to the large variation range of the scale of 3D Gaussians, large-sized Gaussians that often span the foreground and background lead to jagged boundaries of segmented objects. To this end, we propose an online boundary trimming method, GaussianTrimmer, which is an efficient and plug-and-play post-processing method capable of trimming coarse boundaries for existing 3D Gaussian segmentation methods. Our method consists of two core steps: 1. Generating uniformly and well-covered virtual cameras; 2. Trimming Gaussian at the primitive level based on 2D segmentation results on virtual cameras. Extensive quantitative and qualitative experiments demonstrate that our method can improve the segmentation quality of existing 3D Gaussian segmentation methods as a plug-and-play method.
% 随着三维高斯在三维场景表示的广泛应用，基于三维高斯的三维场景分割方法也逐渐兴起。然而，现有的三维高斯分割方法基本上是以高斯基元为单位进行分割，由于三维高斯的scale的变化范围很大，经常有横跨前景和背景的大尺寸高斯导致被分割物体的边缘有毛刺现象。为了解决这个问题，我们提出了一种在线修理边缘的方法，GaussianTrimmer。我们的方法是一种高效且即插即用的后处理方法，包含两个核心步骤：1，生成均匀且良好覆盖的虚拟相机；2，基于虚拟相机上的二维分割结果对高斯边缘进行基元级别的修剪。大量定量和定性的实验证明，我们的方法作为即插即用的方法能提升现有三维高斯分割方法的分割质量。
\end{abstract}


\section{Introduction}
\label{sec:intro}

\begin{figure}[t]
  \centering
  \setlength{\abovecaptionskip}{0.3cm}
  \includegraphics[width=1\linewidth]{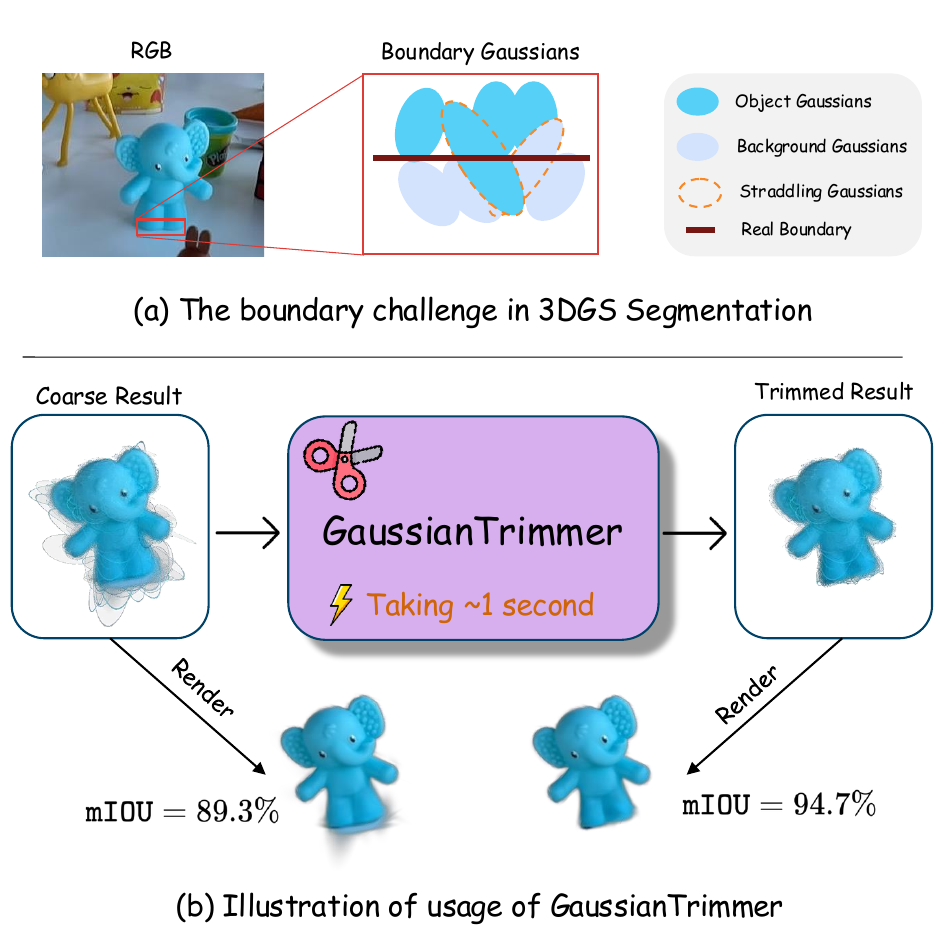}
  \vspace{-0.5cm} 
  \caption{(a) Straddling Gaussians make it difficult to achieve precise segmentation boundaries; (b) Our \textbf{GaussianTrimmer} effectively trims these boundaries for improved segmentation quality within only approximately 1 second latency.}
  \label{fig:teaser}
\end{figure}

In recent years, 3D Gaussian splatting (3DGS)~\cite{3dgs} has gradually occupied an important position in 3D scene representation due to its high-quality rendering and real-time rendering speed. With the widespread application of 3D Gaussians, 3D scene segmentation methods~\cite{GauGroup, SAGA, choi2024click,shen2024flashsplat, zhu2025rethinking,SAGD,qin2024langsplat, wu2024opengaussian,li2025instancegaussian, zhang2025cob, liao2025zero,zhao2025isegman, liao2024clipgs} based on 3D Gaussians have also gradually emerged. However, most existing 3DGS-based segmentation methods suffer from two key limitations: (1) they treat individual Gaussian primitives as the atomic unit for segmentation, and (2) boundary-straddling Gaussians always lie on object boundaries (as illustrated in Fig.~\ref{fig:teaser} (a)). These straddling Gaussians, which overlap both foreground and background regions, hinder the attainment of sharp segmentation boundaries, often resulting in jagged boundaries. Such irregularities not only degrade quantitative metrics like mIoU and mAcc but also compromise visual quality. These shortcomings severely constrain the utility of 3DGS in downstream tasks, including 3D scene editing (e.g., object relocation or compositing) and robotic applications (e.g., grasping simulations). The jagged boundaries of the segmented objects are refered to as ``\textit{boundary challenge}''.

To resolve the boundary challenge, we tackle it from the perspective of decomposing straddling primitives, by further splitting each straddling Gaussian into two smaller Gaussians that respectively belong to the object and the background, thereby achieving training-free trimming of 3D Gaussians. Inspired by SAGD~\cite{SAGD}, we adapt an online method for splitting straddling Gaussians, by decomposing the 2D covariance of the projected Gaussians using segmentation boundaries from the 2D view to guide the decomposition ratio of the 3D covariance, and separately computing the new positions corresponding to the decomposed Gaussians. Unlike previous methods, we treat boundary trimming as a form of post-processing and refinement of the segmentation results (as shown in Fig.~\ref{fig:teaser} (b)). To this end, we devise a boundary trimming strategy based on virtual camera planning, thereby enabling segmentation refinement with extremely low latency (approximately 1s).
% 近年来，由于其优质的渲染质量和实时的渲染速度，三维高斯泼溅逐渐在三维场景表示中占据重要地位。随着三维高斯表示的广泛应用，基于三维高斯的三维场景分割方法也逐渐兴起。然而，现有的三维高斯分割方法基本上是以高斯基元为单位进行分割，并且他们对单个高斯基元通常不会进一步分割。由于三维高斯的scale的变化范围很大，经常有横跨前景和背景的大尺寸高斯导致被分割物体的边缘有毛刺现象，影响分割质量。为了解决这个问题，我们提出了一种在线修理边缘的方法，GaussianTrimmer。我们的方法是一种高效且即插即用的后处理方法，包含两个核心步骤：1，生成均匀且良好覆盖的虚拟相机；2，基于虚拟相机上的二维分割结果对高斯边缘进行基元级别的修剪。大量定量和定性的实验证明，我们的方法作为即插即用的方法能提升现有三维高斯分割方法的分割质量。
Our key contributions can be summarized as:
\begin{itemize}
    \item We propose GaussianTrimmer, an online boundary trimming method for 3DGS segmentation, which effectively addresses the boundary challenge within approximately 1 second latency.
    \item We design an effective Virtual Camera Planning (VCP) module for more precise 3D Gaussian decomposition through the 2D masks.
    \item Extensive experiments demonstrate our GaussianTrimmer can work as a plug-and-play module to    improve the quantitative and qualitative results.
\end{itemize}

\section{Related Works}
\label{Related}
\subsection{3D Neural Scene Segmentation}
Since 3DGS was established as both training-efficient and capable of real-time rendering, a range of exceptional 3DGS-specific segmentation methods have emerged ~\cite{GauGroup, SAGA, choi2024click,shen2024flashsplat, zhu2025rethinking,SAGD,qin2024langsplat, wu2024opengaussian,li2025instancegaussian, zhang2025cob, liao2025zero,zhao2025isegman, liao2024clipgs}. Specifically, GaussianGrouping~\cite{GauGroup} links masks across views using a tracker, then applies joint learning to optimize reconstruction and identity encodings simultaneously. ClickGaussian~\cite{choi2024click}, a post-processing approach, enhances pre-trained 3D Gaussians with two-level granularity features and introduces Global Feature-guided Learning to mitigate mask inconsistencies across views. Certain studies~\cite{SAGD,zhang2025cob} focus on refining boundaries in 3DGS segmentation. Meanwhile, Flashsplat~\cite{shen2024flashsplat} accelerates the optimization process by approximately 50 times, though it still requires 30 seconds for optimization, classifying it as an offline method. 

\subsection{Boundary Refinement}
Decomposing 3D primitives has been explored in various contexts. In 3D reconstruction, methods like SAGD~\cite{SAGD} decompose Gaussians based on 2D segmentation boundaries to improve object delineation. Other works have focused on decomposing meshes or point clouds to enhance segmentation quality. Our approach builds upon these ideas by introducing an online boundary trimming method specifically designed for 3DGS segmentation, addressing the challenges posed by straddling Gaussians. COB-GS~\cite{zhang2025cob} also addresses boundary issues but focuses on a different aspect of Gaussian representation. In contrast, our method emphasizes efficient and effective trimming of boundaries as a post-processing step, enhancing existing segmentation results with minimal latency.

Unlike previous methods that require retraining or complex optimization, our GaussianTrimmer operates as a plug-and-play solution, making it easy to integrate into existing 3DGS segmentation pipelines. Moreover, our method focuses on real-time performance, ensuring that boundary refinement can be achieved with minimal computational overhead.

\section{Method}
\begin{figure*}[ht]
  \centering
    \setlength{\abovecaptionskip}{0.3cm}

  \includegraphics[width=1\linewidth]{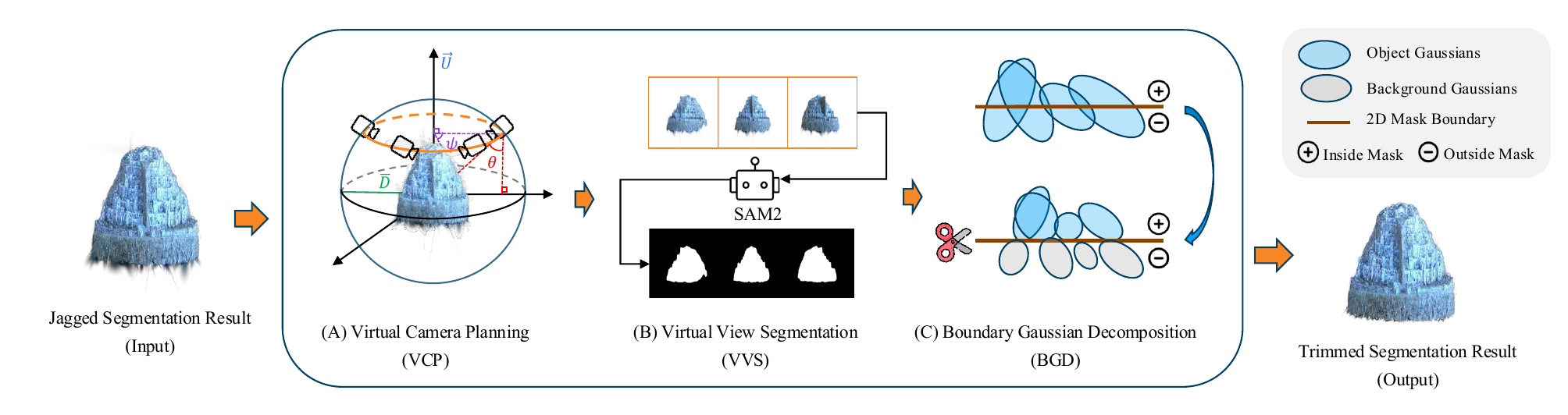}
  \caption{
    \textbf{GaussianTrimmer Pipeline }: Guided by user interaction, Virtual Camera Generation (VCG) module generates smooth and object-center virtual views, followed by Rendering-Tracking-Pruning (RTP) loop on the generated virtual views to identify Gaussians belonging to the target, which are represented as a Gaussian-level Boolean 3D mask.}
  \label{fig: framework}
\end{figure*}

In this section, we introduce GaussianTrimmer, a method for online trimming boundaries in 3DGS segmentation. As illustrated in Fig.~\ref{fig: framework}, GaussianTrimmer is a post-processing module designed to refine the segmentation results from existing 3DGS segmentation methods. The input to GaussianTrimmer is a jagged segmentation result, and the output result has smoother boundaries. This step can be formulized as follows:
\begin{equation}
    \hat{\mathbf{\Theta}_{\mathcal{O}}} = \texttt{GauTrimmer}(\mathbf{\Theta}_{\mathcal{O}}; \mathbf{\Theta}, \vec{U}),
\end{equation}
where $\mathbf{\Theta} \in \mathbb{R}^{N \times C}$ represents total Gaussians of the scene, $\mathbf{\Theta}_{\mathcal{O}} \in \mathbb{R}^{K\times C}$ is a subset of $\mathbf{\Theta}$, denoting the jagged coarse segmentation result obtained from an existing 3DGS segmentation method, and \(\vec{U}\) denotes the up vector used for virtual camera orientation. The output $\hat{\mathbf{\Theta}_{\mathcal{O}}} \in \mathbb{R}^{K \times C}$ is the refined segmentation result with improved boundary quality. Note that $\mathbf{\Theta}$ is optional and can be used for background augmentation during virtual view segmentation.
% 如图所示，我们的GaussianTrimmer是一个对已分割的三维目标进行后处理的模块。GaussianTrimmer的输入是jagged segmentation result，而输出是一个边缘更加平滑的分割结果。

\subsection{Virtual Camera Planning (VCP)}
\begin{figure}[th]
  \centering
  \setlength{\abovecaptionskip}{0.3cm}
  \includegraphics[width=1\linewidth]{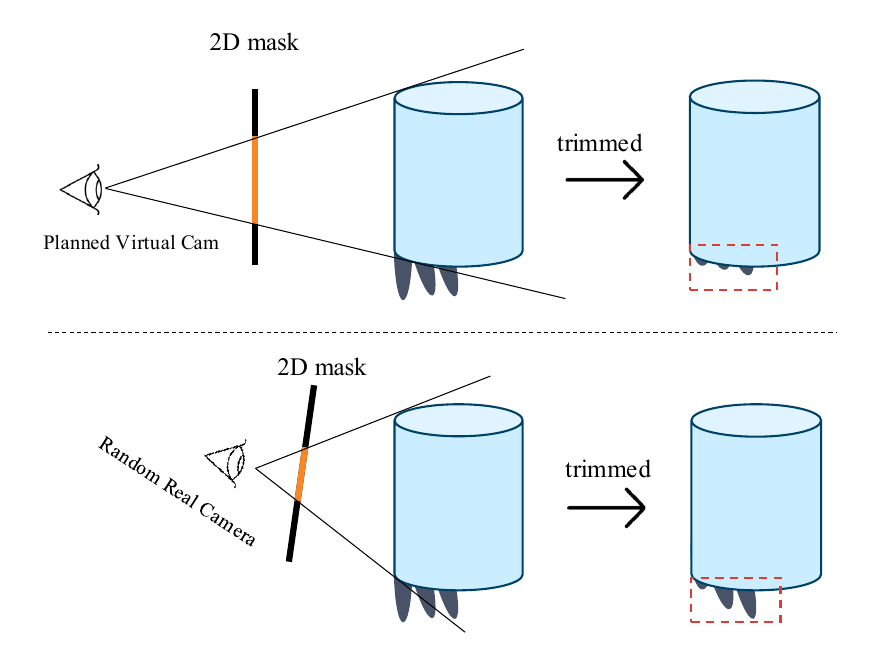}
  \vspace{-0.5cm} 
  \caption{\textbf{Our virtual cameras vs. real cameras}. Our planned virtual cameras (top row) provide better object coverage and alignment compared to real cameras (bottom row), leading to more accurate trimming results.}
  \label{fig:vcam}
\end{figure}

Corresponding to Fig.~\ref{fig: framework} (A), the first step of our pipeline is Virtual Camera Planning (VCP). Different from previous boundary refinement methods that utilize 2D masks from training views as cues, we employ virtual views that offer better alignment and more comprehensive coverage (see Fig.~\ref{fig:vcam}). 
% 与之前的boundary refinement方法不同，我们的方法没有在训练视角上使用2D Mask作为线索，我们采用了对齐性更好、覆盖更全的虚拟视角。
The VCP module generates a set of virtual cameras $\mathbf{c} = \{c_0, c_1, ..., c_{n-1}\}$ that uniformly cover the segmented object $\mathbf{\Theta}_{\mathcal{O}}$. To achieve this, we adopt a \textit{spherical coordinate system} as follows:
First, we determine the centroid of the object, which serves as the center of the spherical coordinate system. The centroid $L$ is computed using the positions of the Gaussians in the segmented object $\mathbf{\Theta}_{\mathcal{O}}$ as
% 为了实现这一点，我们采用了球形坐标系来规划一条围绕目标的虚拟相机轨迹。首先，我们确定出目标的质心，然后以此为球形坐标系的圆心，质心公式求解如下：
\begin{equation}
    L = \frac{1}{|\boldsymbol{p}|} \sum_{\mathbf{x} \in \boldsymbol{p}} \mathbf{x},
    \label{eq:locate_object}
\end{equation} 
where $\boldsymbol{p} = \{\mathbf{x}_i | \mathbf{x}_i \in \mathbf{\Theta}_{\mathcal{O}}\}$ represents the set of positions of Gaussians in the segmented object, and $|\boldsymbol{p}|$ is the total number of these Gaussians.
Next, we determine suitable distance $r$ from the centroid $L$ to position the virtual cameras. Let $\{l, w, h\}$ denote length, width and height of $\mathbf{\Theta}_{\mathcal{O}}$. We set the distance $r$ as
\begin{equation}
    r = \texttt{max}(\lambda_1 \cdot \texttt{max}(l, w), \lambda_2 \cdot h),
    \label{eq:camera_distance}
\end{equation}
where $\lambda_1$ and $\lambda_2$ are hyperparameters to control the distance and keep the whole object visible in virtual views. In our setting, we set $\lambda_1 = 20$ and $\lambda_2 = 6$.

Then, we construct a spherical coordinate system $\mathcal{Y}$ with $L$ as the center, $r$ as the radius and $\vec{U}$ as the up axis. In $\mathcal{Y}$, we can place the virtual cameras based on yaw and pitch angles $\{\psi, \theta\}$. The $\vec{U}$ can be obtained by manual calibration or normal vector estimation of the ground plane via RANSAC~\cite{fischler1981RANSAC}. We fix the pitch angle $\theta$ to ensure all virtual cameras lie on the same plane, and then uniformly sample the yaw angle $\psi$ within a specific range. For each camera position $P_i$, it can be represented as:
\begin{equation}
    P_i = (\psi_i, \theta, r).
    \label{eq:camera_position}
\end{equation}
Then, we orient each virtual camera to face the target centroid $L$, generating the final camera poses. Finally, we set the intrinsic parameters for each virtual camera to ensure that its field of view covers the entire target.
For intrinsic matrix generation, we set:\begin{equation}
\mathbf{K} =
\begin{pmatrix}
\dfrac{w}{2 \tan\left( \frac{FOV}{2} \right)} & 0 & \dfrac{w}{2} \\[12pt]
0 & \dfrac{h}{2 \tan\left( \frac{FOV}{2} \right)} & \dfrac{h}{2} \\[12pt]
0 & 0 & 1
\end{pmatrix}
    \label{eq:intrinsic}
\end{equation}
where $w$ and $h$ are the width and height of the rendered image, respectively. $FOV$ is the field of view, which is set to $60^\circ$ in our experiments. 
% 接下来，我们以L为球心，r为半径，构建一个球坐标系Y。在Y中，我们只需根据俯仰角和偏航角即可确定出虚拟相机的位置。我们固定俯仰角使得所有虚拟相机处于同一个平面，然后在特定角度范围内均匀采样偏航角。对于每个相机位置P_i，可以表示为：
% 然后，我们将每个虚拟相机都朝向目标质心L，生成最终的相机位姿。
% 最后，我们为每个虚拟相机设置内参，使其视野能够覆盖整个目标。

\subsection{Virtual View Segmentation (VVS)}
After planning the virtual cameras, we proceed to the Virtual View Segmentation (VVS) step, as depicted in Fig.~\ref{fig: framework} (B). In this step, we render the scene from the perspective of each virtual camera to generate a series of virtual views. Each virtual view captures a 2D projection of the 3D scene, allowing us to leverage 2D segmentation techniques for boundary refinement. The Segment Anything (SAM) series~\cite{kirillov2023segment,ravi2024sam2}, especially SAM2, is employed to perform segmentation on these virtual views. SAM2 is chosen for its robustness and accuracy in cross-view tracking. For cross-view tracking, we design two strategies to better suit the Gaussian boundary processing: 1) automatic mask generation; 2) background augmentation.
% 在虚拟相机规划好以后，我们渲染出对应的虚拟视图，然后把多视图当做视频的连续帧来处理。
% 我们使用2D视频跟踪算法SAM2来完成这一操作。
% 对于cross-view tracking的使用，我们设计了两个策略使其更适合用于高斯边界处理：1，自动生成初始mask；2，背景信息增强。

\noindent \textbf{Automatic Mask Generation.} To bootstrap the segmentation process in SAM2, we automatically generate an initial mask for the first virtual view. Subsequently, we leverage SAM2's cross-view tracking capabilities to propagate the segmentation to the remaining virtual views.
For the initial mask generation, we render the coarse segmentation output $\mathbf{\Theta}_{\mathcal{O}}$. Since the background in the first virtual view $I_0$ is blank, the bounding box of the target object can be directly extracted. This bounding box is then provided as a prompt to SAM2 to produce a refined initial mask using
 \begin{equation}
    m_0=\texttt{SAM2}(I_0; \texttt{box}(I_0)).
    \label{eq:initial_mask}
\end{equation}

\noindent \textbf{Background Augmentation.} Due to the blank background in the virtual views, SAM2 may mistakenly classify the jagged boundaries as part of the foreground object. To address this issue, we employ a background augmentation strategy. Specifically, we augment the coarse segmentation result $\mathbf{\Theta}_{\mathcal{O}}$ by incorporating additional background Gaussians to obtain $\mathbf{\Theta}_{\mathcal{O}}^{\prime}$ as:
 \begin{equation}
    \mathbf{\Theta}_{\mathcal{O}}^{\prime}=\mathbf{\Theta}_{\mathcal{O}} \cup \mathbf{\Theta}_{n},
    \label{eq:BA}
\end{equation}
where $\mathbf{\Theta}_{n}$ represents the set of neighboring Gaussians of $\mathbf{\Theta}_{\mathcal{O}}$. Then we render $\mathbf{\Theta}_{\mathcal{O}}^{\prime}$ to obtain virtual views for 2D masks.
% 由于背景几乎是blank，在初始分割时，SAM2容易认为毛刺的边界也会属于前景物体的一部分，这导致GaussianTrimmer无法有效地修剪边界。
% 为了解决这个问题，我们采用了背景增强的方法。具体来说，我们对粗糙分割结果进行背景扩充，包括更多的背景高斯参与虚拟视图渲染，以此得到更准确的二维分割结果，这个能指导GaussianTrimmer进行有效的修剪。

\subsection{Boundary Gaussian Decomposition (BGD)}
BGD module is the core module of GaussianTrimmer. As shown in Fig.~\ref{fig: framework} (C), the BGD module operates on the straddling Gaussians located at the boundaries. By decomposing overly long Gaussians into two shorter ones, it effectively addresses the boundary jaggedness issue. Thus, the goal of the BGD module is to: (1) detect these straddling Gaussians and (2) perform decomposition processing on them.
% Boundary Gaussian Decomposition (BGD)是GaussianTrimmer的核心模块。如图~\ref{fig: framework} (C)所示，BGD模块作用于位于边界出的straddling Gaussians，通过将过长的高斯分解成 2 个短高斯来出来边界毛刺问题。所以BGD模块的目标是识别出这些边界高斯，并对其进行分解处理。

\noindent\textbf{Boundary Gaussians Detection}. To efficiently identify the straddling boundary Gaussians, we employ a reverse $\alpha$-blending strategy. Specifically, we trace back from a single pixel in the 2D virtual view to the set of Gaussians that contribute to the color value of that pixel. Since the number of Gaussians contributing to each pixel's rendering varies, we implement a top-$k$ selection scheme, selecting the top-$k$ Gaussians based on their contribution to the pixel's color rendering. By constructing a one-to-$k$ pixel-to-Gaussian index mapping, we can efficiently index all straddling Gaussians for the contour points of the 2D mask in the virtual view. Thus, we can quickly identify all straddling Gaussians $\mathbf{\Theta}_{b}$ that may cause jagged boundaries, which is a small subset compared to the entire set of object Gaussians $\mathbf{\Theta}_{\mathcal{O}}$.
% 为了更快并且更准确地找到边界高斯，我们采用了逆向渲染策略。具体说，我们根据二维视图的单个像素索引出决定这个像素颜色值的高斯集合。由于渲染每个像素参与的高斯数量是不等的，我们设定了top-k选取方案，按照像素渲染的贡献度选取前k个高斯。通过构建了 1 对k的像素-高斯索引映射，我们只需通过二维mask的contour点即可索引出该视点下的所有straddling Gaussians。因此，我们能快速锁定所有的会引起毛刺的高斯，相比于物体高斯而言，是一个很小的子集。

\begin{figure}[th]
  \centering
  \setlength{\abovecaptionskip}{0.3cm}
  \includegraphics[width=0.8\linewidth]{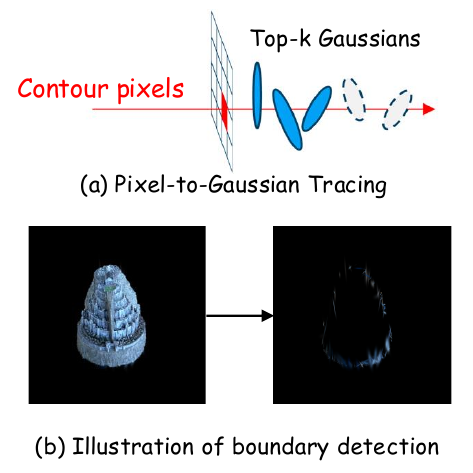}
  \vspace{-0.5cm} 
  \caption{(a) Illustration of pixel-to-Gaussian mapping; (b) Ilustration of boundary Gaussian detection.}
  \label{fig:boundary}
\end{figure}

\noindent\textbf{Gaussian Decomposition}. Once the straddling Gaussians $\mathbf{\Theta}_{b}$ are identified, we proceed to decompose them into two shorter Gaussians. For each straddling Gaussian, we first determine its principal axis based on its covariance matrix. We then split the Gaussian along this axis at its centroid, creating two new Gaussians with adjusted means and covariances. The weights and colors of the new Gaussians are inherited from the original Gaussian. This decomposition effectively reduces the length of the straddling Gaussians, leading to smoother boundaries in the segmentation result. For a single Gaussian decomposition whose long axis and position are denoted as $\{l, \mathbf{x}\}$, we follow the SAGD~\cite{SAGD} to figure out one of the new Gaussians ($\{l^{\prime}, \mathbf{x}^{\prime}\}$) as
\begin{equation}
l^{\prime}=\lambda l,
\end{equation}
\begin{equation}
\mathbf{x}^{\prime}=\mathbf{x}+\frac{1}{2}\left(l-\lambda l\right) \mathbf{e},
\end{equation}
where $\lambda$ is the decomposition ratio, and $\mathbf{e}$ is the unit vector along the principal axis.
Finally, we prune the Gaussians that are not consistently identified as part of the target object across multiple views, resulting in a refined segmentation with smoother boundaries.

\section{Experiments}
\subsection{Experimental Setup}
To demonstrate that our GaussianTrimmer is an effective post-processing method for 3DGS segmentation results, we primarily evaluate the improvement in segmentation quality brought by GaussianTrimmer to existing 3DGS segmentation methods in our quantitative experiments. We conduct comparative experiments with existing 3DGS segmentation methods on multiple benchmark datasets and perform both quantitative and qualitative analyses. For quantification, we utilize the NVOS dataset~\cite{ren2022NOVS}, which is derived from the LLFF~\cite{mildenhall2019llff} dataset and provides ground truth masks with precise object edges. For qualitative evaluation, we employ scenes from various datasets, including IN2N~\cite{haque2023instruct}, Mipnerf-360~\cite{barron2022mip}, PKU-DyMVHuman~\cite{zheng2024pku}, and LERF-Mask~\cite{kerr2023lerf}.
% 为了证明GaussianTrimmer在是一种有效的3DGS分割结果后优化方法，我们在定量实验中主要评估GaussianTrimmer对现有3DGS分割方法分割质量的提升效果。我们在多个基准数据集上与现有的3DGS分割方法进行了对比实验，并进行了定量和定性的分析。

% 我们先用现有3DGS分割方法得到一个初始的三维分割结果，然后用GaussianTrimmer对初始分割结果进行边缘修剪，最后评估修剪前后的分割质量提升效果。
\noindent\textbf{Implementation Details}. We first obtain an initial 3DGS segmentation result using existing 3DGS segmentation methods, then apply GaussianTrimmer for boundary trimming on the initial segmentation result, and finally evaluate the improvement in segmentation quality before and after trimming. In all quantitative experiments, we set the number of virtual cameras to 5 for each target object.

\begin{figure*}[ht]
  \centering
    \setlength{\abovecaptionskip}{0.3cm}

  \includegraphics[width=1\linewidth]{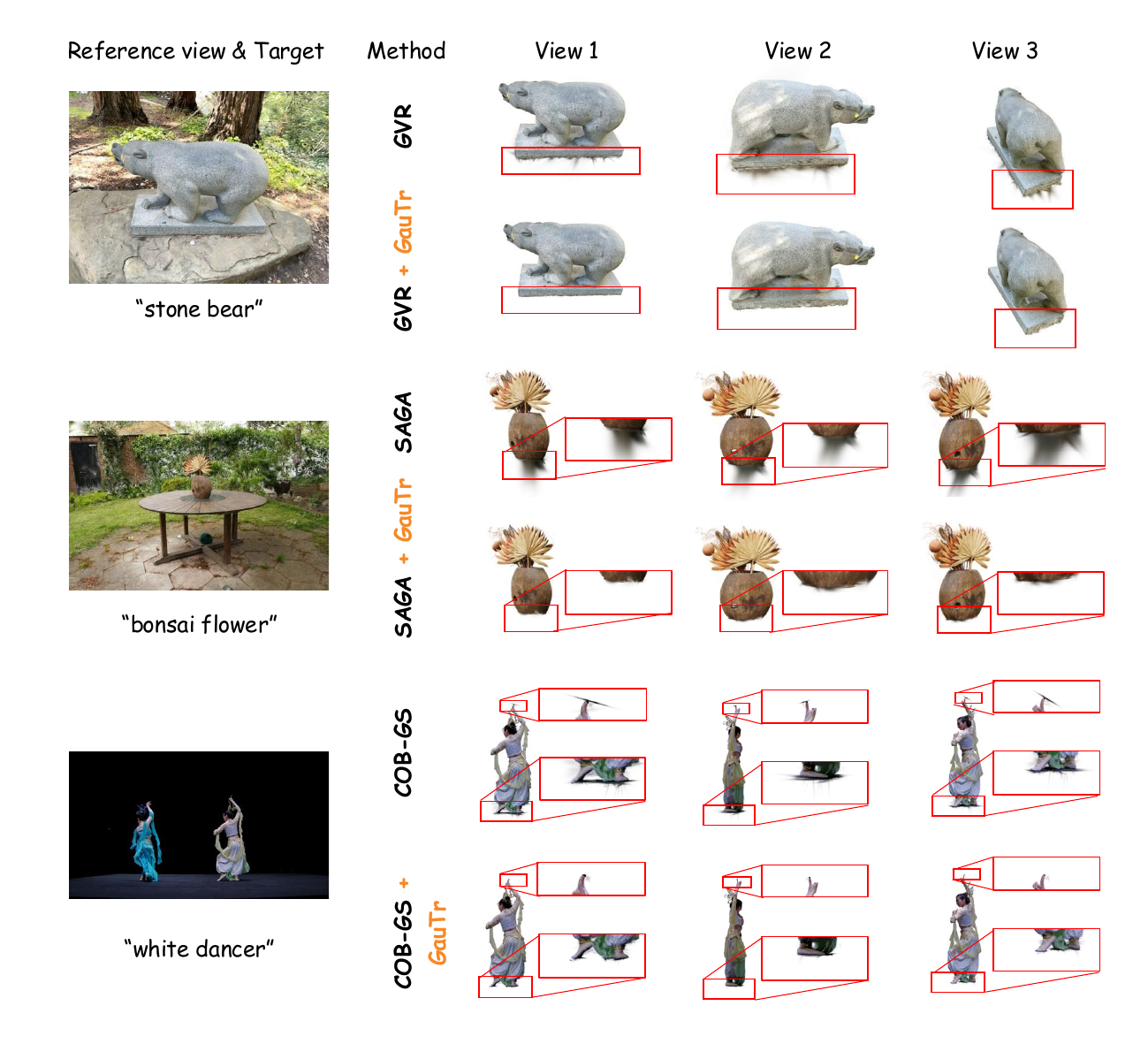}
      \vspace{-0.5cm}
  \caption{
    \textbf{Qualitative Results.} This figure showcases the direct improvement effects of GaussianTrimmer on the segmentation results of existing methods. From a visual perspective, GaussianTrimmer effectively refines jagged edges and enhances the clarity of segmentation boundaries.
    %此图展示了GaussianTrimmer对现有方法分割结果的直接提升效果。从可视化角度来看，GaussianTrimmer能够有效改善毛刺边缘，提高分割边界的清晰度。 
    }
    \vspace{-0.3cm}
  \label{fig: comparison}
\end{figure*}

\subsection{Quantitative Results}
We evaluate the effectiveness of our GaussianTrimmer on the NVOS dataset~\cite{ren2022NOVS} by applying it to enhance the segmentation results of several state-of-the-art 3DGS segmentation methods, including SA3D~\cite{cen2023SA3D}, OmniSeg3D~\cite{ying2024omniseg3d}, SAGA~\cite{SAGA}, FlashSplat~\cite{shen2024flashsplat}, SAGD~\cite{SAGD}, and COB-GS~\cite{zhang2025cob}. The quantitative results are summarized in \cref{tab:nvos1}. Our method consistently improves the mIoU and mAcc metrics across all evaluated methods, demonstrating its effectiveness as a plug-and-play post-processing technique for enhancing 3DGS segmentation quality.

\begin{table}[htbp]
  \centering
  \caption{Quantitative segmentation results on NVOS dataset.}
  \label{tab:nvos1}
  \small
  \begin{tabular}{lccc}
  \toprule
  Method     & mIoU (\%) & mAcc (\%) \\
  \midrule
  SA3D~\cite{cen2023SA3D} & 90.3 & 98.2 \\
  + \texttt{GaussianTrimmer} & 91.9 \textcolor{ForestGreen}{(+1.6)} & 98.5 \textcolor{ForestGreen}{(+0.3)} \\
  \midrule
  OmniSeg3D~\cite{ying2024omniseg3d} & 91.7 & 98.4 \\
  + \texttt{GaussianTrimmer} & 92.4 \textcolor{ForestGreen}{(+0.7)} & 98.7 \textcolor{ForestGreen}{(+0.3)} \\
  \midrule
  SAGA~\cite{SAGA} & 90.9 & 98.3 \\
  + \texttt{GaussianTrimmer} & 92.1 \textcolor{ForestGreen}{(+1.2)} & 98.5 \textcolor{ForestGreen}{(+0.2)} \\
  \midrule
  FlashSplat~\cite{shen2024flashsplat} & 91.8 & 98.6 \\
  + \texttt{GaussianTrimmer} & 92.2 \textcolor{ForestGreen}{(+0.4)} & \textbf{98.7} \textcolor{ForestGreen}{(+0.1)} \\
  \midrule
  SAGD~\cite{SAGD} & 90.4 & 98.2 \\
  + \texttt{GaussianTrimmer} & 92.0 \textcolor{ForestGreen}{(+1.6)} & 98.5 \textcolor{ForestGreen}{(+0.3)} \\
  \midrule
  COB-GS~\cite{zhang2025cob} & 92.1 & 98.6 \\
  + \texttt{GaussianTrimmer} & \textbf{92.5} \textcolor{ForestGreen}{(+0.4)} & \textbf{98.7} \textcolor{ForestGreen}{(+0.1)} \\
  \bottomrule
  \end{tabular}
  \vspace{-5pt}
\end{table}

\begin{table}[htbp]
    \centering
    \caption{Quantitative visual results on NVOS dataset.}
    \label{tab:nvos2}
    \small
    \begin{tabular}{l @{\hskip 10pt} c @{\hskip 5pt} c @{\hskip 5pt} c}
        \toprule
        \multirow{3}{*}{Method} & \multicolumn{3}{c}{CLIP-IQA~\cite{wang2022exploring} (\%) $\uparrow$} \\ \cmidrule(lr){2-4}
                                      & \scriptsize{\begin{tabular}[c]{@{}c@{}}\textit{Clear} / \textit{Unclear} \\ \textit{Boundary}\end{tabular}} & \scriptsize{\begin{tabular}[c]{@{}c@{}}\textit{Smooth} / \textit{Noisy} \\ \textit{Boundary}\end{tabular}} & \scriptsize{\begin{tabular}[c]{@{}c@{}}\textit{Complete} / \textit{Mutilated} \\ \textit{Object}\end{tabular}} \\ \midrule
        SAGD~\cite{SAGD}      & 0.621 & 0.631 & 0.788 \\
        + \texttt{GauTrimmer}      & 0.688 & 0.738 & 0.835 \\
        \midrule
        COB-GS~\cite{zhang2025cob}    & 0.682 & 0.731 & 0.859 \\
        + \texttt{GauTrimmer}      & \textbf{0.695} & \textbf{0.742} & \textbf{0.867} \\
        \bottomrule
    \end{tabular}
    \vspace{-3pt}
\end{table}

% \begin{table}[t]
%     \centering
%     \caption{Ablation results on NVOS dataset. BAGS indicates the boundary-adaptive Gaussian splitting; BGTR indicates the boundary-guided texture restoration; RAEM indicates robustness against erroneous masks.}
%     \label{tab:a2}
%     \renewcommand{\arraystretch}{0.9}
%     \begin{tabular}{ccc|cc}
%     \toprule
%     \multicolumn{3}{c|}{Component} & \multicolumn{2}{c}{Performance} \\
%     \midrule
%     BAGS & BGTR & RAEM & mIoU (\%) & mAcc (\%)  \\
%     \midrule
%     &  &  & 91.2 & 98.3 \\
%     \checkmark &  &  & 91.9 & 98.5 \\
%     \ding{51} & \ding{51} &  & 91.9 & 98.4 \\
%     \ding{51} & \ding{51} & \ding{51} & \textbf{92.1} & \textbf{98.6} \\
%     \bottomrule
%     \end{tabular}
%     \vspace{-10pt}
% \end{table}
Moveover, we follow Zhang et al.~\cite{zhang2025cob} to assess the visual quality of segmentation results before and after applying GaussianTrimmer using the CLIP-IQA~\cite{wang2022exploring} metric, which evaluates the clarity of object boundaries and the completeness of segmented objects. As shown in \cref{tab:nvos2}, GaussianTrimmer significantly enhances the visual quality of segmentation results, yielding higher CLIP-IQA scores across all evaluated aspects.

\subsection{Qualitative Results}
We further demonstrate the effectiveness of GaussianTrimmer through qualitative comparisons on various real-world scenes from multiple datasets, including IN2N~\cite{haque2023instruct}, Mipnerf-360~\cite{barron2022mip}, PKU-DyMVHuman~\cite{zheng2024pku}. As illustrated in \cref{fig: comparison}, GaussianTrimmer effectively refines the segmentation boundaries and recovers missing object parts, leading to more accurate and visually appealing segmentation results. These qualitative results further validate the capability of GaussianTrimmer to enhance 3DGS segmentation quality across diverse scenarios.

\subsection{Ablation Study}
We conduct ablation studies on three aspects: virtual cameras, background augmentation, and the number of virtual viewpoints. Firstly, we replace the virtual viewpoints with real training viewpoints to verify the effectiveness of the VCP module. Then we disable the background augmentation module to validate its contribution to segmentation quality improvement. Finally, we evaluate the performance and latency of GaussianTrimmer under different numbers of virtual viewpoints to balance segmentation quality and computational overhead (see~\cref{tab:n_varying}).
% 为了更好地理解GaussianTrimmer中各个关键组件的作用，我们对背景增强、虚拟相机、虚拟视点数量等三个方面进行了消融实验。首先我们用真实的训练视点来代替虚拟视点，以验证VCP模块的有效性。然后我们关闭背景增强模块，以验证背景增强对分割质量提升的贡献。最后，我们在不同虚拟视点数量下评估了GaussianTrimmer的性能和延迟，以权衡分割质量和计算开销之间的关系。
\begin{table}[htbp]
  \centering
  \caption{Ablation study on NVOS dataset taking SAGA as baseline.}
  \label{tab:ablation}
  \small
  \begin{tabular}{lccc}
  \toprule
  Method     & mIoU (\%) & mAcc (\%) \\
  \midrule
  SAGA~\cite{SAGA} (baseline) & 90.9 & 98.3 \\
  \midrule
  No VCP & 91.3 & 98.3 \\
  No Bg. Augmentation & 91.5 & 98.4 \\
  No VCP and BA & 91.3 & 98.3 \\
  Full GaussianTrimmer & \textbf{92.1} \textcolor{ForestGreen}{(+1.2)} & \textbf{98.5} \textcolor{ForestGreen}{(+0.2)} \\
  \bottomrule
  \end{tabular}
  \vspace{-0.7cm}
\end{table}

\begin{table}[th]
\centering
\caption{ablation on varying virtual camera number.
}
\setlength{\tabcolsep}{2mm}
\renewcommand{\arraystretch}{1.15} % Adjust row spacing
\begin{tabular}{lcccccccc}
\noalign{\hrule height 1.2pt}
$n=$ & 1 & 2 & 3 & 5 & 7 & 10\\
\midrule
Latency(ms) & 256 & 409 & 521 & \textbf{766} & 1037 & 1566 \\
mIOU(\%) Gain & 0.8 & 0.9 & 1.0 & \textbf{1.2} & 1.2 & 1.2 \\ 
\noalign{\hrule height 1.2pt}
\end{tabular} 
%}

\label{tab:n_varying}
 \vspace{-0.1cm}
\end{table}

\section{Conclusion}
In this paper, we have presented GaussianTrimmer, an efficient and plug-and-play online boundary trimming method designed to enhance the segmentation quality of existing 3D Gaussian Splatting (3DGS) segmentation methods. By generating uniformly distributed virtual cameras and leveraging 2D segmentation results to trim straddling Gaussians, our approach effectively addresses the boundary challenge inherent in 3DGS segmentation. Extensive experiments on benchmark datasets demonstrate that GaussianTrimmer significantly improves segmentation accuracy, particularly in refining object boundaries. Our method serves as a valuable post-processing step that can be seamlessly integrated into various 3DGS segmentation pipelines, paving the way for more precise and reliable 3D scene understanding.

\bibliographystyle{IEEEbib}
\bibliography{icme2026references}

@String(CVPR  = {CVPR})

@String(ICCV  = {ICCV})

@String(ECCV  = {ECCV})

@String(TOG   = {ACM TOG})

@String(AAAI = {AAAI})

@article{3dgs,
   author = {Kerbl, Bernhard and Kopanas, Georgios and Leimkühler, Thomas and Drettakis, George},
   title = {3d gaussian splatting for real-time radiance field rendering},
   journal = TOG,
   volume = {42},
   number = {4},
   pages = {1-14},
   ISSN = {0730-0301},
   year = {2023},
   type = {Journal Article}
}

@inproceedings{GauGroup,
  title={Gaussian grouping: Segment and edit anything in 3d scenes},
  author={Ye, Mingqiao and Danelljan, Martin and Yu, Fisher and Ke, Lei},
  booktitle=ECCV,
  pages={162--179},
  year={2024},
  organization={Springer}
}

@article{SAGA,
  title={Segment any 3d gaussians},
  author={Cen, Jiazhong and Fang, Jiemin and Yang, Chen and Xie, Lingxi and Zhang, Xiaopeng and Shen, Wei and Tian, Qi},
  journal={arXiv preprint arXiv:2312.00860},
  year={2023}
}

@inproceedings{shen2024flashsplat,
  title={Flashsplat: 2d to 3d gaussian splatting segmentation solved optimally},
  author={Shen, Qiuhong and Yang, Xingyi and Wang, Xinchao},
  booktitle=ECCV,
  pages={456--472},
  year={2024},
  organization={Springer}
}

@inproceedings{choi2024click,
  title={Click-gaussian: Interactive segmentation to any 3d gaussians},
  author={Choi, Seokhun and Song, Hyeonseop and Kim, Jaechul and Kim, Taehyeong and Do, Hoseok},
  booktitle=ECCV,
  pages={289--305},
  year={2024},
  organization={Springer}
}

@article{zhu2025rethinking,
  title={Rethinking End-to-End 2D to 3D Scene Segmentation in Gaussian Splatting},
  author={Zhu, Runsong and Qiu, Shi and Liu, Zhengzhe and Hui, Ka-Hei and Wu, Qianyi and Heng, Pheng-Ann and Fu, Chi-Wing},
  journal={arXiv preprint arXiv:2503.14029},
  year={2025}
}

@article{SAGD,
  title={SAGD: Boundary-enhanced segment anything in 3D Gaussian via Gaussian decomposition},
  author={Hu, Xu and Wang, Yuxi and Fan, Lue and Fan, Junsong and Peng, Junran and Lei, Zhen and Li, Qing and Zhang, Zhaoxiang},
  journal={arXiv preprint arXiv:2401.17857},
  year={2024}
}

@article{zhang2025cob,
  title={COB-GS: Clear Object Boundaries in 3DGS Segmentation Based on Boundary-Adaptive Gaussian Splitting},
  author={Zhang, Jiaxin and Jiang, Junjun and Chen, Youyu and Jiang, Kui and Liu, Xianming},
  journal={arXiv preprint arXiv:2503.19443},
  year={2025}
}

@inproceedings{qin2024langsplat,
  title={Langsplat: 3d language gaussian splatting},
  author={Qin, Minghan and Li, Wanhua and Zhou, Jiawei and Wang, Haoqian and Pfister, Hanspeter},
  booktitle={Proceedings of the IEEE/CVF Conference on Computer Vision and Pattern Recognition},
  pages={20051--20060},
  year={2024}
}

@inproceedings{li2025instancegaussian,
  title={Instancegaussian: Appearance-semantic joint gaussian representation for 3d instance-level perception},
  author={Li, Haijie and Wu, Yanmin and Meng, Jiarui and Gao, Qiankun and Zhang, Zhiyao and Wang, Ronggang and Zhang, Jian},
  booktitle={Proceedings of the Computer Vision and Pattern Recognition Conference},
  pages={14078--14088},
  year={2025}
}

@article{wu2024opengaussian,
  title={Opengaussian: Towards point-level 3d gaussian-based open vocabulary understanding},
  author={Wu, Yanmin and Meng, Jiarui and Li, Haijie and Wu, Chenming and Shi, Yahao and Cheng, Xinhua and Zhao, Chen and Feng, Haocheng and Ding, Errui and Wang, Jingdong and others},
  journal={arXiv preprint arXiv:2406.02058},
  year={2024}
}

@article{liao2025zero,
  title={Zero-Shot Visual Grounding in 3D Gaussians via View Retrieval},
  author={Liao, Liwei and Li, Xufeng and Zheng, Xiaoyun and Liu, Boning and Gao, Feng and Wang, Ronggang},
  journal={arXiv preprint arXiv:2509.15871},
  year={2025}
}

@inproceedings{zhao2025isegman,
  title={iSegMan: Interactive Segment-and-Manipulate 3D Gaussians},
  author={Zhao, Yian and Xu, Wanshi and Zheng, Ruochong and Qiao, Pengchong and Liu, Chang and Chen, Jie},
  booktitle={Proceedings of the Computer Vision and Pattern Recognition Conference},
  pages={661--670},
  year={2025}
}

@article{liao2024clipgs,
  title={Clip-gs: Clip-informed gaussian splatting for real-time and view-consistent 3d semantic understanding},
  author={Liao, Guibiao and Li, Jiankun and Bao, Zhenyu and Ye, Xiaoqing and Wang, Jingdong and Li, Qing and Liu, Kanglin},
  journal={arXiv preprint arXiv:2404.14249},
  year={2024}
}

@article{fischler1981RANSAC,
  title={Random sample consensus: a paradigm for model fitting with applications to image analysis and automated cartography},
  author={Fischler, Martin A and Bolles, Robert C},
  journal={Communications of the ACM},
  volume={24},
  number={6},
  pages={381--395},
  year={1981},
  publisher={ACM New York, NY, USA}
}

@inproceedings{kirillov2023segment,
  title={Segment anything},
  author={Kirillov, Alexander and Mintun, Eric and Ravi, Nikhila and Mao, Hanzi and Rolland, Chloe and Gustafson, Laura and Xiao, Tete and Whitehead, Spencer and Berg, Alexander C and Lo, Wan-Yen and others},
  booktitle=ICCV,
  pages={4015--4026},
  year={2023}
}

@article{ravi2024sam2,
  title={SAM 2: Segment Anything in Images and Videos},
  author={Ravi, Nikhila and Gabeur, Valentin and Hu, Yuan-Ting and Hu, Ronghang and Ryali, Chaitanya and Ma, Tengyu and Khedr, Haitham and R{\"a}dle, Roman and Rolland, Chloe and Gustafson, Laura and Mintun, Eric and Pan, Junting and Alwala, Kalyan Vasudev and Carion, Nicolas and Wu, Chao-Yuan and Girshick, Ross and Doll{\'a}r, Piotr and Feichtenhofer, Christoph},
  journal={arXiv preprint arXiv:2408.00714},
  url={https://arxiv.org/abs/2408.00714},
  year={2024}
}

@inproceedings{ren2022NOVS,
  title={Neural volumetric object selection},
  author={Ren, Zhongzheng and Agarwala, Aseem and Russell, Bryan and Schwing, Alexander G and Wang, Oliver},
  booktitle={Proceedings of the IEEE/CVF Conference on Computer Vision and Pattern Recognition},
  pages={6133--6142},
  year={2022}
}

@inproceedings{kerr2023lerf,
  title={Lerf: Language embedded radiance fields},
  author={Kerr, Justin and Kim, Chung Min and Goldberg, Ken and Kanazawa, Angjoo and Tancik, Matthew},
  booktitle={Proceedings of the IEEE/CVF International Conference on Computer Vision},
  pages={19729--19739},
  year={2023}
}

@inproceedings{zheng2024pku,
  title={PKU-DyMVHumans: A Multi-View Video Benchmark for High-Fidelity Dynamic Human Modeling},
  author={Zheng, Xiaoyun and Liao, Liwei and Li, Xufeng and Jiao, Jianbo and Wang, Rongjie and Gao, Feng and Wang, Shiqi and Wang, Ronggang},
  booktitle={Proceedings of the IEEE/CVF Conference on Computer Vision and Pattern Recognition},
  pages={22530--22540},
  year={2024}
}

@article{mildenhall2019llff,
  title={Local light field fusion: Practical view synthesis with prescriptive sampling guidelines},
  author={Mildenhall, Ben and Srinivasan, Pratul P and Ortiz-Cayon, Rodrigo and Kalantari, Nima Khademi and Ramamoorthi, Ravi and Ng, Ren and Kar, Abhishek},
  journal={ACM Transactions on Graphics (ToG)},
  volume={38},
  number={4},
  pages={1--14},
  year={2019},
  publisher={ACM New York, NY, USA}
}

@inproceedings{barron2022mip,
  title={Mip-nerf 360: Unbounded anti-aliased neural radiance fields},
  author={Barron, Jonathan T and Mildenhall, Ben and Verbin, Dor and Srinivasan, Pratul P and Hedman, Peter},
  booktitle=CVPR,
  pages={5470--5479},
  year={2022}
}

@inproceedings{ying2024omniseg3d,
  title={Omniseg3d: Omniversal 3d segmentation via hierarchical contrastive learning},
  author={Ying, Haiyang and Yin, Yixuan and Zhang, Jinzhi and Wang, Fan and Yu, Tao and Huang, Ruqi and Fang, Lu},
  booktitle={Proceedings of the IEEE/CVF Conference on Computer Vision and Pattern Recognition},
  pages={20612--20622},
  year={2024}
}

@article{cen2023SA3D,
  title={Segment anything in 3d with nerfs},
  author={Cen, Jiazhong and Zhou, Zanwei and Fang, Jiemin and Shen, Wei and Xie, Lingxi and Jiang, Dongsheng and Zhang, Xiaopeng and Tian, Qi and others},
  journal={Advances in Neural Information Processing Systems},
  volume={36},
  pages={25971--25990},
  year={2023}
}

@inproceedings{wang2022exploring,
    title = {{Exploring CLIP for Assessing the Look and Feel of Images}},
    author = {Wang, Jianyi and Chan, Kelvin CK and Loy, Chen Change},
    booktitle = {AAAI},
    year = {2023}
}

@inproceedings{haque2023instruct,
  title={Instruct-nerf2nerf: Editing 3d scenes with instructions},
  author={Haque, Ayaan and Tancik, Matthew and Efros, Alexei A and Holynski, Aleksander and Kanazawa, Angjoo},
  booktitle={Proceedings of the IEEE/CVF international conference on computer vision},
  pages={19740--19750},
  year={2023}
}

\end{document}